\documentclass[a4paper, 10pt, conference]{IEEEtran}      

\usepackage{graphics} 
\usepackage{epsfig} 
\usepackage{mathptmx} 
\usepackage{times} 
\usepackage{amsmath} 
\usepackage{amssymb}  
\usepackage{parskip}
\let\labelindent\relax
\usepackage{enumitem}
\usepackage{caption}
\usepackage{xcolor}
\usepackage{booktabs}
\usepackage{subcaption}
\usepackage{cite}
\def\BibTeX{{\rm B\kern-.05em{\sc i\kern-.025em b}\kern-.08em
    T\kern-.1667em\lower.7ex\hbox{E}\kern-.125emX}}

\usepackage{hyperref}
\usepackage[capitalise]{cleveref}

\title{\LARGE \bf
Local Planner Bench: Benchmarking for Local Motion Planning}

\author{Max Spahn$^{*}$, Chadi Salmi$^{*}$, Javier Alonso-Mora$^{*}$
\thanks{*This research was supported by Ahold Delhaize and ERC. All content represents the opinion of the authors, which is not necessarily shared or endorsed by their respective employers and/or sponsors.}
\thanks{The authors are with the Department of Cognitive Robotics, Delft University of Technology, 2628 CD, Delft, The Netherlands
       {\tt\small \{m.spahn, c.salmi,j.alonsomora\}@tudelft.nl}}
}

\begin{document}

\bstctlcite{IEEEexample:BSTcontrol}

\newcommand{\lpb}{localPlannerBench}

\maketitle
\thispagestyle{empty}
\pagestyle{empty}

\newlist{todolist}{itemize}{2}
\setlist[todolist]{label=$\square$}


\newcommand{\missing}[1]{\textbf{{\color{red}{{#1}}}}} 

\begin{abstract}

Local motion planning is a heavily researched topic in the field of robotics with many promising algorithms being published every year. 
However, it is difficult and time-consuming to
compare different methods in the field.
In this paper, we present \lpb, a new benchmarking suite that allows quick and seamless comparison between local motion planning algorithms.
The key focus of the project lies in the extensibility of the environment and the simulation cases. 
Out-of-the-box, localPlannerBench already supports many simulation cases ranging from a simple 2D point
mass to full-fledged 3D 7DoF manipulators, and it's straightforward to add your own custom robot using a URDF file. 
A post-processor is built-in that can be extended with custom metrics and plots. 
To integrate your own motion planner, simply create a wrapper that derives from the provided base class. 
Ultimately we aim to improve the reproducibility of local motion planning algorithms and encourage standardized open-source comparison. \\
\\
Local Motion Planning, Benchmark, Collision Avoidance
\\
Code: \href{https://github.com/tud-amr/localPlannerBench}{github.com/tud-amr/localPlannerBench}
\footnote{version 1.0.0: \href{https://github.com/tud-amr/localPlannerBench/tree/v.1.0.0}{https://github.com/tud-amr/localPlannerBench/tree/v.1.0.0}}
\end{abstract}

\section{INTRODUCTION}

Robot motion planning in dynamic environments is fundamentally different from motion generation in static environments because
initial plans must be constantly updated in the presence of unforeseen events.
Algorithms that aim to adapt global plans according to dynamic changes in the environment and transfer them into
actions at runtime fall into the category of \textit{local motion planning} or \textit{reactive motion planning}.
As the number of applications (e.g. mobile robots, robotic arms, mobile manipulators) is quite diverse, different methods
have been presented in the last years showing varying performance depending on the scenarios. 
These methods can be roughly divided into four categories:
(a) geometric approaches, such as the potential field method \cite{potential_fields}, reciprocal collision avoidance with velocity obstacles \cite{van2011reciprocal}, Riemannian Motion Policies \cite{rmps} or
optimization fabrics \cite{fabrics,dynamic_fabrics},
(b) optimization-based approaches, such as STOMP \cite{stomp} or model predicitive control \cite{mpc,gompc},
(c) compositon of motion primitives \cite{motion_primitives,motion_primitives_hauser}, and
(d) learning-based approaches \cite{wang_learning_2020,jurgenson_harnessing_2019}.
Due to a steady increase in number of methods and
a tendency to focus on one of the above approaches,
it becomes increasingly difficult and time-consuming to make proper comparisons
among methods.
Additionally, new methods in the field are often not publicly available so comparisons require re-implementation
of these methods. Together with a general bias toward optimizing parameters of a proposed method, 
fair comparisons become more challenging.
To address these issues, we offer \lpb{}, 
a bench-marking tool to allow quick and seamless comparison between different local motion planning algorithms.
Additionally, we promote extendibility to allow users to modify \lpb{} according to their setups.
 
First, we review briefly other benchmarking software with a similar purpose while laying out the differences to our
framework (\cref{sec:state}).
Then, we state the scope of our framework and introduce the approach. (\cref{sec:approach}).
We lay out the general structure of \lpb{} and explain individual components in \cref{sub:code_structure}.
Lastly, we briefly present a use-case where two different planners are compared (\cref{sec:example}).
%
%
\section{RELATED WORK}
\label{sec:state}
As we propose \lpb{} for benchmarking local motion planning algorithms, we briefly revise existing
benchmarks in robotics to highlight differences. Sampling-based motion planning for global motion planning
was first standardized in the Open Motion Planning Library (OMPL) \cite{ompl}. There, many algorithms are implemented and 
accessible to the user. Today, it is used as the backend planning library for many robotics software \cite{moveit,openrave}.
While the benefit of assembling different algorithms in a single library is, still today, highlighted by OMPL, it is not
a benchmark suite on its own. Robowflex is a wrapper around various motion planning libraries that can evaluate different
global motion planning algorithms in isolation \cite{robowflex}. It also includes standardized motion planning
problems to promote comparisons. The goal behind \lpb{} is to provide a similar framework but instead of focusing
on global motion planning methods, which are most valuable in static environments, we focus on dynamic environments and
thus on local motion planning methods.
Some benchmarking frameworks focus purely on mobile robots and/or autonomous driving and provide
benchmarks for local motion planning methods in cluttered and human-shared environments \cite{arena_bench,bench_mr}.
With \lpb{}, we generalize this approach beyond mobile robots by including robotic arms and mobile manipulators.
To make motion planning methods more reproducible and comparable, all of the mentioned works play a central role and
\lpb{} takes this role for local motion planning.
\begin{figure}
    \centering
    \begin{subfigure}[b]{0.40\textwidth}
         \centering
         \includegraphics[width=0.9\linewidth]{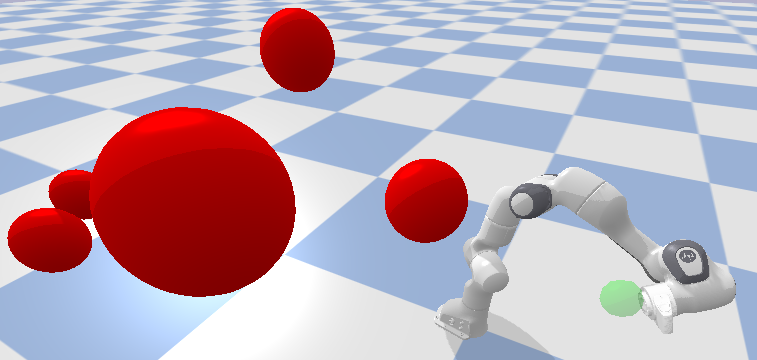}
         \caption{}
         \label{fig:example_env_3d}
     \end{subfigure}
    \begin{subfigure}[b]{0.20\textwidth}
         \centering
         \includegraphics[width=0.9\linewidth]{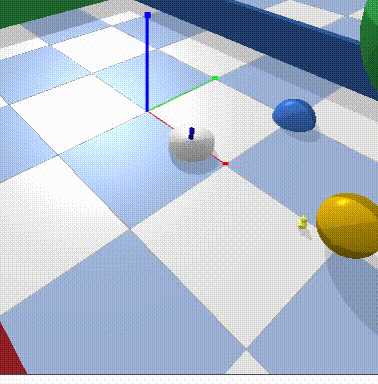}
         \caption{}
         \label{fig:example_env_3d_point}
     \end{subfigure}%
     \begin{subfigure}[b]{0.20\textwidth}
         \centering
         \includegraphics[width=0.9\linewidth]{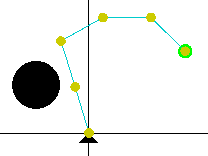}
         \caption{}
         \label{fig:example_env_planar}
     \end{subfigure}
    \caption{Examples of environments (pybullet in \cref{fig:example_env_3d,fig:example_env_3d_point} and simple ODE
    in \cref{fig:example_env_planar}) available in \lpb{}. Goal states are depicted as green circles, and
    obstacles as either red or black depending on the environment.}
    \label{fig:example_environments}
\end{figure}

\section{APPROACH}
\label{sec:approach}

\subsection{Scope}

We define local motion planning as the problem of computing actions for the
robot at runtime, such that the sequence of actions leads to progress in the
fulfillment of the goal while avoiding collisions with its environment. From
this definition, it follows that interactive motion planning, where the robot
is allowed to manipulate objects, is not considered. Moreover, we consider
collision avoidance the only safe way of navigating through the environment. 
While the initial state can be specified by a robot configuration, the goal is
a composition of geometric constraints. Some examples that can be stated as
geometric constraints are, that the end-effector has to be at a defined
position and/or orientation, some joints need to be in a specific position,
some links must align or the robot has to follow a certain trajectory. Although
not supported in this version, the latter could be generated with global
planner. In other words, the meaningful behavior of the robot in the
environment is assumed to be parameterizable by these geometric goals. 

Furthermore, perception pipelines are outside the scope of this work. At this
stage, we assume a working perception pipeline that can detect all obstacles as
they move through the environment.  In the future, we want to integrate more
abstract perception pipelines, e.g. generating occupancy grids based on the
obstacles positions and shapes. Low-level control of the joints of the robot is
also out of the scope of this work. That means, we provide environments with
different control modalities, but the control signal is directly passed to the
environment, so a perfect low-level controller is assumed.

\subsection{Definitions}
\label{sub:definitions}

An \textbf{environment} refers to the physics simulator (or real world) that
produces robot states and obstacle poses and applies actions to the robot. The
\textbf{problem formulation} is defined by the geometric parameterization of
the goal, the robot, the specifications of the obstacles and additional global
environment variables. A \textbf{planner} is a local motion planning method
that computes an action based on the current state of the environment. We
define a \textbf{metric} as a quantitative measure for the performance
evaluation of a planner. A single run of the environment with corresponding
problem formulation and planner is referred to as a \textbf{trial}. A
\textbf{study}, within \lpb{}, refers to a series of trials.

\subsection{Rational}

Our guidelines for developing and maintaining \lpb{} are the following:
\begin{itemize}
    \item{\textbf{Simplicity}} It should be as simple as possible to run a trial/study.
    \item{\textbf{Modularity}} It should be straightforward to add a planner, metric or problem formulation.
    \item{\textbf{Specificity}} It should focus on local motion planning and local motion planning only.
\end{itemize}

In the following, we derive design choices based on those motivations.

\subsection{Environment} 

The environment in \lpb{} is not specific to a single physics simulator and
should even be compatible with the real world through e.g. a ROS bridge
(\textit{modularity}). However, at the time of writing, two physics simulators
are officially supported (see \cref{fig:example_environments}), planar
environments are based on a simple ODE-simulation and pybullet
\cite{coumans2021} is used for more complex 3D environments. In the pybullet
environments we use the unified robot description format (URDF) that allows to
import new robots simply by this robot description.
To achieve a clean separation between the simulation environment and planning
algorithm, we opted to design an Open-AI gym \cite{openaigym} environment
wrapper for the different robot types and control modalities that we support
(\textit{Simplicity, Modularity}). Pybullet is not considered the most accurate
physics engine, but it is much easier to install compared to competing
simulators like ISAAC-Gym \cite{IsaacGym} or MuJoCo \cite{todorov2012mujoco}
(\textit{Simplicity}). Besides, as low-level control and interactive motion
planning are outside of our scope, the accuracy of PyBullet is considered
sufficient (\textit{Specificity}).

\begin{figure*}
    \centering
    \includegraphics[width=1.0\linewidth]{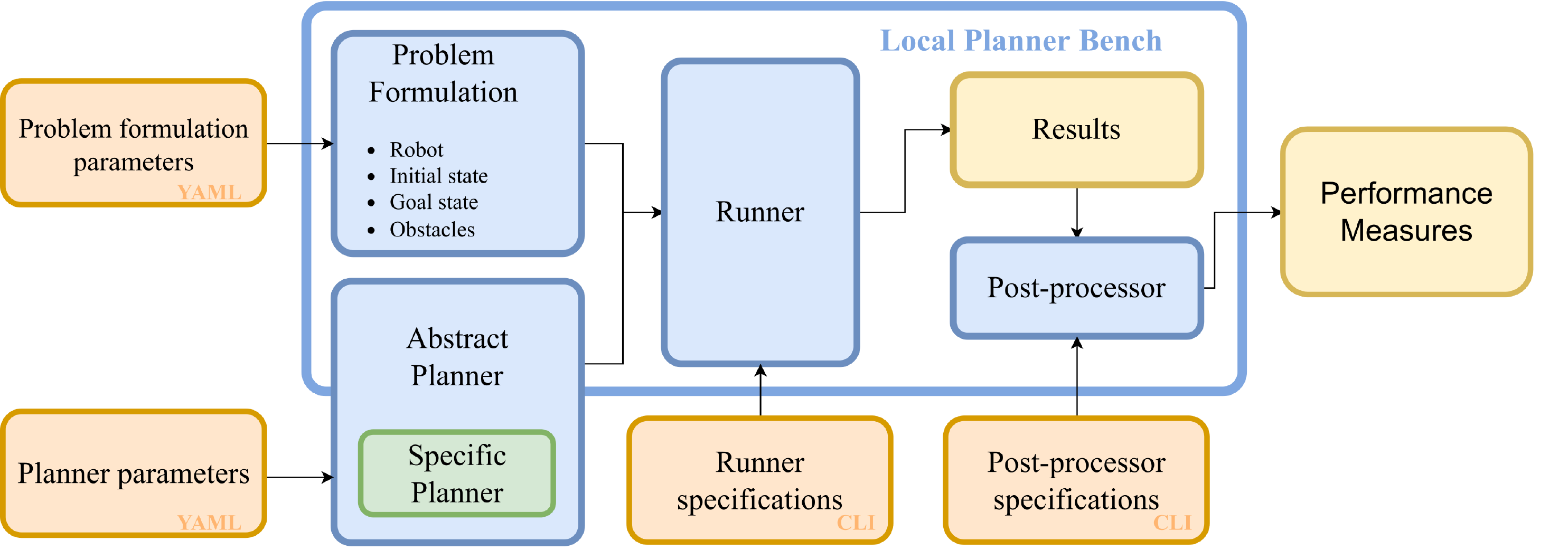}
    \caption{Code structure of \textbf{localPlannerBench}. Blue components are provided, while the orange components should be passed along at run-time. A specific planner (green) could be implemented by the user, and the outputs are shown as yellow blocks.}
    \label{fig:code_structure}
\end{figure*}

\subsection{Code structure}
\label{sub:code_structure}
The code in \lpb{} is split up into components as can be seen in
\cref{fig:code_structure}. Each component has a unique and clearly defined
responsibility (\textit{Modularity}). The parameters specifying specific
configurations of these components are as much as possible defined outside of
the code itself using \textsc{yaml-files}. When possible, the code structure
reflects the definitions proposed in \cref{sub:definitions}:
\begin{itemize}
    \item{The \textsc{Problem}} class corresponds to the problem formulation
      and requires its own configuration file to read in the requested problem
      formulation. Additionally, the component supports randomization of the
      initial configuration, goal state, and obstacle states. The limits of the
      randomization should also be defined within the configuration file.
    \item The \textsc{AbstractPlanner} is a blueprint for a planner class. When
      a user creates a new planner, she should derive from this provided base
      class, so that the planner will automatically be registered as a valid
      planner to the rest of the package. A planner created using this
      component is parameterized by the corresponding configuration file.
    \item{The \textsc{runner}} executable is the entry point when running a
      trial or study i.e. it is the executable that the user interacts with.
      The component contains the code for the actual execution loops over the
      Open-AI gym simulation environments and saves the results afterward.
    \item{The \textsc{post-processor}} executable processes the results based
      on user specified metrics and plot characteristics. Like the runner, it
      is an executable that the user interacts with.
    
\end{itemize}

\subsection{Usage}

There are currently two general possible usages of \lpb{}: (a) access an open
implementation of a local motion planner and its tuned parameters by an expert
or (b) compare your own local motion planner against baselines that are already
implemented in \lpb{}

\subsection{Running a study} \label{sub:run}

\lpb{} provides a minimal command-line interface (CLI) that processes the
configuration files described in the previous section. The CLI calls the
\textit{runner} that runs a study with the given parameters. Specifically, the
user specifies the location of the configurations files for the problem
formulation and the planner. Additionally, flags and arguments for the study
can be speficied through the CLI. For example, the user could specify the
location of the output files, whether the simulation should be rendered or not,
how many trials should be run, etc. Finally, all relevant information to
reproduce a study is saved in a time-stamped results folder. An overview of
arguments can be found in Table \ref{table:runner_args}.

\begin{table}
\centering{
\begin{tabular}{l c l}
\toprule
Argument & & Explanation \\
\midrule
caseSetup & : & Path to problem formulation file. \\
planners & : & List of paths to planner configuration files. \\
res-folder & : & Path to store the result data. \\
render & : & Display GUI. \\
numberRuns & : & How many trials to run. \\
random-goal & : & Should the goal be random? \\
random-init & : & Should the initial configuration be random? \\
random-obst & : & Should the obstacles be random? \\
\bottomrule
\end{tabular}
}
\caption{List of CLI arguments for the runner executable  \label{table:runner_args}}
\end{table}

\subsection{Postprocessing a study} \label{sub:postprocess}

To promote better reproducibility, running, and post-processing a study is
completely separated. Specifically, the output of a study is the raw data that
was directly accessible during the execution. That is: the joint position, the
joint velocities, the taken actions, the positions of the links of the robot,
the position of obstacles, and the goal at each time step. During
post-processing, the raw data is evaluated according to user metrics and is
potentially plotted. Several performance metrics are implemented, see the
repository for a full list. However, additional metrics could be added using a
similar approach as for the planner. An overview of post-processor arguments
can be found in Table \ref{table:post_processor_args}.

\begin{table} 
\centering{
\begin{tabular}{l c l}
\toprule
Argument & & Explanation \\
\midrule
expFolder & : & Path to the results of a study. \\
kpis & : & List of all kpis/metrics to be evaluated. \\
plot & : & Auto-generate plot from results. \\
series & : & Are you evaluating a single trial or multiple? \\
compare & : & Are you comparing planners? \\
\bottomrule
\end{tabular}
}
\caption{CLI arguments for the post\_processor executable \label{table:post_processor_args}}
\end{table}

\section{EXAMPLE: Optimization fabrics vs MPC}
\label{sec:example}
\begin{figure}
    \begin{subfigure}{0.95\linewidth}
        \centering
        \includegraphics[angle=-90,width=0.9\textwidth]{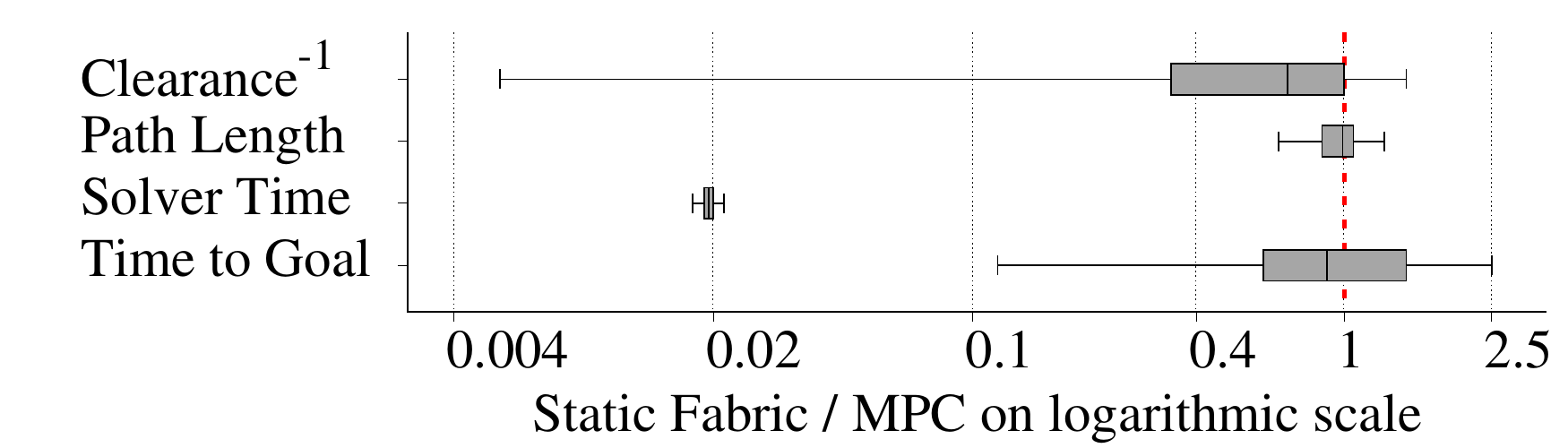}
        \caption{}
        \label{fig:success_plot}
    \end{subfigure}
    \begin{subfigure}{0.95\linewidth}
        \centering
        \includegraphics[angle=-90,width=0.9\textwidth]{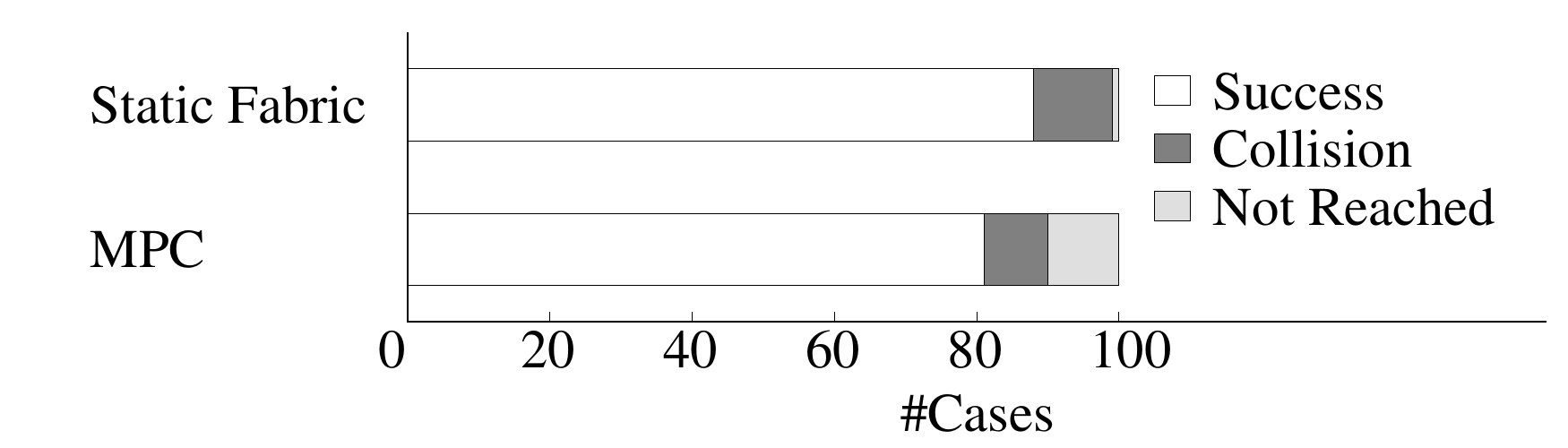}
        \caption{}
        \label{fig:success_plot}
    \end{subfigure}
    \caption{
        Exemplary comparison between two different local motion planning algorithms in \lpb{}.
    }
    \label{fig:example_fabrics_mpc}
\end{figure}

An example of how \lpb{} can be used to compare the two methods, Static Fabrics
and Model Predictive Control (MPC), is found in \cite{fabrics}. The goal of the
comparison was to investigate if the significant reduction in solver time of
Static Fabrics impacts other metrics like path length, clearance or time to
goal. Using the runner, a study was run for a hundred trials, including both
planners. Afterwards, the plots in Figure \ref{fig:example_fabrics_mpc} were
generated using the post-processor.

\section{CONCLUSION} \label{sec:conclusion}

As the field of local motion planning matures, the need for standardized
reproducible benchmarks increases. To make the life of a researcher a little
easier, we've created \lpb. This framework is capable of simulating local
motion planning problems, for many different robots, modalities, and
environments. Furthermore, \lpb{} is fully open source and designed with
simplicity, modularity, and specificacy in mind, enabling users to easily
modify \lpb{} to fit their unique use cases as long as they fit within the
scope as what we define as local motion planning. In the future, we hope to
extend \lpb{} in several ways: integrating basic global planning methods to
provide reference trajectories for local planning, adding artificial
uncertainty to the states and observations to help make the studies more
representative of real scenarios, adding support for observations in the form
of occupancy maps, semantic maps, or even direct camera or lidar data to
accommodate new directions in local motion planning. \lpb{} has already been
proven to be useful for academic comparison and evaluations \cite{fabrics}, and
we hope it will gain futher attention in this field of local motion planning.
\bibliographystyle{IEEEtran}
\bibliography{bib.bib}
\end{document}